\RequirePackage{fix-cm}

\documentclass{svClass}          
\usepackage[margin=1in]{geometry}
\smartqed  
\usepackage{graphicx}
\usepackage{subcaption}
\usepackage{longtable} 

\usepackage[square,numbers]{natbib}
\makeatletter
\newcommand{\BeginOneColumn}{\if@twocolumn\onecolumn\gdef\EndOneColumn{\twocolumn}\else\gdef\EndOneColumn{}\fi}
\makeatother
\begin{document}

\title{Deep Dive into Identity and Cloth-Aware Modeling: Review of Techniques and Datasets for 2D-to-3D Human Reconstruction}

\author{%
\centering
  \textbf{Gulraiz Khan}, \textbf{Kevin Pimbblet}, \textbf{Kenneth Y. Wertheim}\\[4pt]
  {\small\itshape Centre of Excellence for Data Science, Artificial Intelligence and Modelling (DAIM),}\\
  {\small\itshape University of Hull, Hull, UK}\\[4pt]
  {\small G.Khan-2022@hull.ac.uk, K.Pimbblet@hull.ac.uk, K.Y.Wertheim@hull.ac.uk}}

\institute{Gulraiz Khan  \at
			  University of Hull \\
              \email{g.khan-2022@hull.ac.uk}
}

\titlerunning{Identity- and Cloth-Aware 2D-to-3D Human Reconstruction: A Review}
\authorrunning{G. Khan, K. Pimbblet, K. Y. Wertheim}

\maketitle
\pagestyle{plain} 

\begin{abstract}
3D human reconstruction from 2D images has become a growing field and has attracted researchers from many disciplines. 3D reconstruction has wide applications in multimedia, gaming, medical, and education. Advances in facial landmark detection, pose estimation, and parametric mesh generation have improved 3D reconstruction. Achieving high performance and fidelity in 3D human reconstruction in the presence of occlusion requires advanced methods. These methods include 3D Gaussian Splatting, and diffusion-based priors. Recently, vision-language models (VLMs) are beginning to supervise and verify reconstruction through natural-language feedback rather than purely geometric losses. This survey provides a comprehensive, pipeline-level review of identity- and cloth-aware 3D human reconstruction. In this review, we traced the evolution from parametric mesh models (e.g., SMPL) and implicit and voxel functions to VLM-based 3D reconstruction engines. To the best of our knowledge, no prior surveys have taken this systematic approach. \\
We followed a systematic four-stage review methodology. (1) Searching and filtering the 92 systems across six categories: face- and body-level feature prior extraction (age, gender, pose, and landmarks), identity-aware reconstruction, cloth-aware modeling, generative and splatting-based methods, VLM-based reconstruction, and benchmark datasets for 3D human reconstruction. (2) Comparison based on fidelity, real-time applicability, and hardware constraints. (3) Trade-off analysis based on fidelity versus real-time ability, identity preservation versus generalization. (4) Finally, we identify six open challenges that must be addressed in the field of 3D human reconstruction for real-world applicability of 3D human reconstruction systems.


\end{abstract}

\section{Introduction}
\label{intro}
\begin{figure*}
\begin{subfigure}{.5\textwidth}
  \centering
  \includegraphics[width=.8\linewidth]{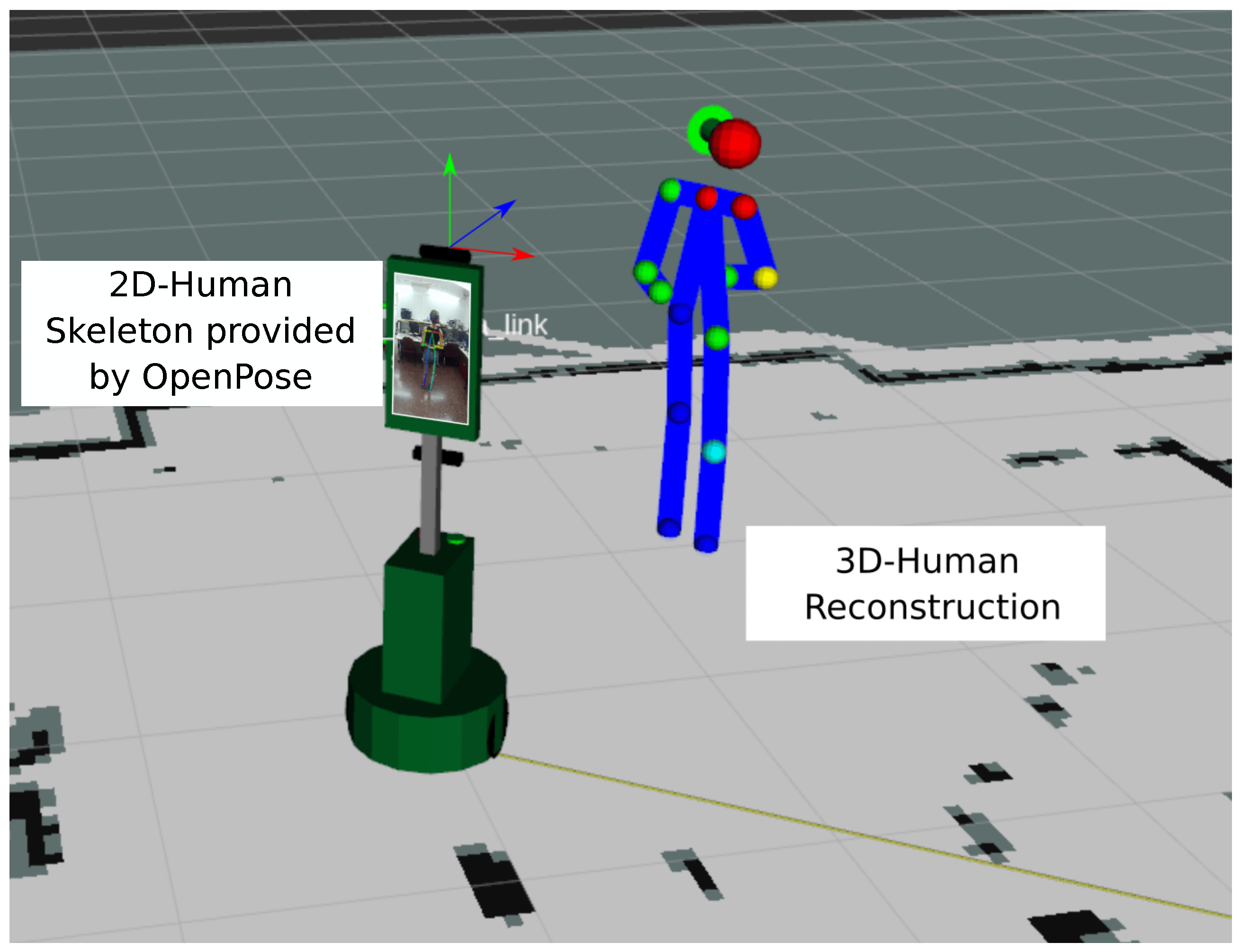}
  \caption{3D Human Reconstruction in Games}
  \label{fig:sfig1}
\end{subfigure}%
\begin{subfigure}{.5\textwidth}
  \centering
  \includegraphics[width=.6\linewidth]{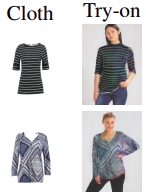}
  \caption{3D Human Reconstruction in try-on systems}
  \label{fig:sfig2}
\end{subfigure}
\begin{subfigure}{1\textwidth}
  \centering
  \includegraphics[width=.8\linewidth]{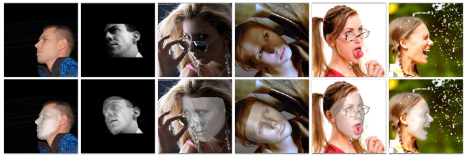}
  \caption{3D Human Reconstruction in Multi-media}
  \label{fig:sfig2}
\end{subfigure}
\caption{Applications of 3D Human Reconstruction in Different Industries}
\label{apps}
\end{figure*}

3D reconstruction plays a pivotal role in computer animation and machine vision. A comprehensive 3D model accurately defines the object's shape, including its length, width, and volume. These detailed 3D models find broad applications in animation, gaming, virtual try-on, medical monitoring, and the entertainment industry (as shown in Figure \ref{apps}). Following are some key applications of identity-aware 3D human modeling.\\

\begin{enumerate}
  \item Gaming and 3D Animation: 3D models are widely used in different human character-based games and animated clips including movies and short videos. 
  Direct transformation of real human subjects into 3D models enhances realism, interactivity, and engagement in games and videos \cite{1intro}.
  \item Virtual Cloth and Makeup Try-On Systems: Identity and cloth-aware 3D Modeling systems can be utilized by clothing and makeup brands to provide a virtual wearing experience to their customers. With just a click, customers can visualize how products will be in reality, helping them make informed buying decisions.\cite{2intro,3intro}.
  \item Healthcare system: 3D modeling can be extensively employed for training and simulating medical procedures. Students and surgeons can perform rehearsal tests on 3D models of their patients. 
  \item 3D Data Creation: Along with real-world physical applications, 3D reconstruction of 2D humans can help in creating 3D data for different human face and body-related machine learning applications. 
\end{enumerate}

A full human primarily consists of many different components including face, head, neck, upper torso, arms and legs. For 3D reconstruction, we are reviewing the full human as two substantial components including the human face and body. We first discussed different face-level features including age, gender, and fiducial points for face-level 3D modeling. Following this, we reviewed the use of body skeleton points to capture human posture. Finally, all face and body-based 3D modeling systems are reviewed based on extracted features. \\
Reconstructing a 2D human image into a 3D model or avatar is known as 2D to 3D reconstruction. This has been achieved by using a variety of CNN-based approaches that leverage human position and movement characteristics. The problem of 3D reconstruction in these approaches is often viewed as a regression and auto-regression problem.\\
Different studies provide similar comparisons of different 3D reconstruction systems. The review provided by Chen et al. \cite{gaussiansplatting_survey}, focuses on 3D Gaussian Splatting pipelines for 3D reconstruction. Another study \cite{wang2024survey} focused on avatar generation using different methods, and \cite{gu20253d} discusses neural radiance field-based avatar reconstruction. These surveys focus specifically on 3D human reconstruction as a standalone task without connecting it back to identity-level features like age, gender, body pose, and facial landmarks. Adding these cues can help in preserving subject's identity. This survey, however, tries to fill the gap in previous reviews by providing a full pipeline review, from reviewing high-level facial features and traditional mesh reconstruction to Gaussian Splatting and diffusion-based avatar models. \\
Alongside the review, we connect it to the proposed EgoTokens, which reconstructs SMPL-X parameters from a single image. This system uses a vision-language model to reconstruct a 3D human mesh as discrete tokens. We note how the system addresses some of the surveyed limitations.\\
We are reviewing a novel aspect of 3D reconstruction systems that utilizes identity, position and movement features for face and body generation. The major contribution of this article is many-fold. \\
\begin{enumerate}
\item We conducted a structural review in three directions: identity- and cloth-aware 3D human reconstruction, face- and body-level feature extraction, and full unified 3D model generation.
\item  This survey covers two recent paradigm shifts in the field: 3D Gaussian Splatting-based avatar generation and diffusion-based 3D human reconstruction.
\item We also review the emerging trend of vision-language model (VLM)- based human reconstruction. Language reasoning supervises, verifies, and controls the overall 3D reconstruction. To the best of our knowledge, no previous survey in the field covers this direction.
\item We compare methods based on real-time applicability, fidelity, and hardware constraints.
\item We also identify open challenges that limit the real-world deployment of 3D reconstruction systems and future directions. We also connect them to our EgoTokens work as a concrete starting point.
\end{enumerate}

\section{Review Methodology}

\begin{figure}[!ht]
    \centering    \includegraphics[width=0.7\linewidth]{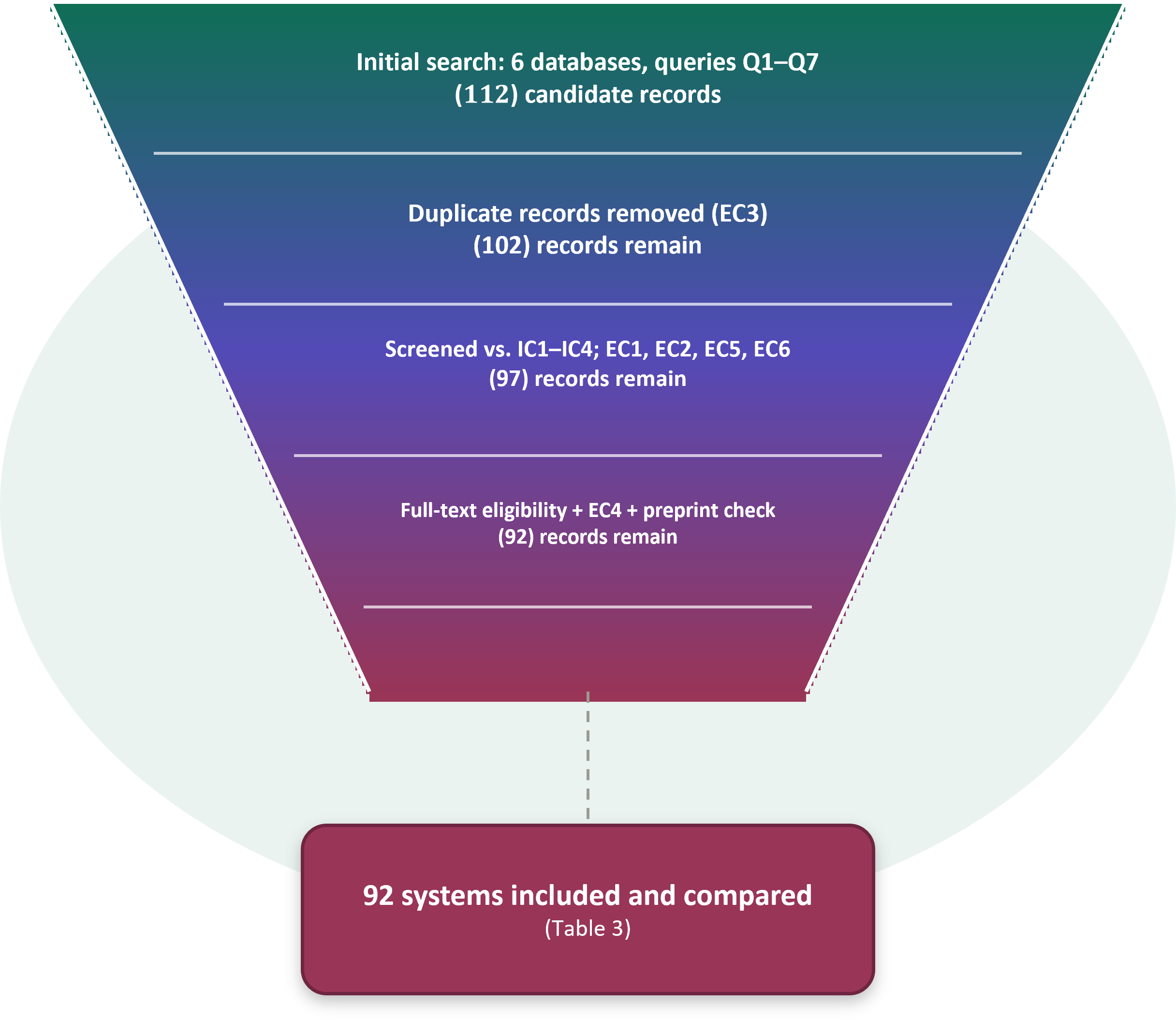}
    \caption{Prisma Selection Funnel. Reducing the number of systems from 112 to 92 included in the review.}
    \label{fig:prisma_funnel}
\end{figure}
\begin{figure*}[!ht]
    \centering
    \begin{subfigure}[b]{0.48\linewidth}
        \includegraphics[width=\linewidth]{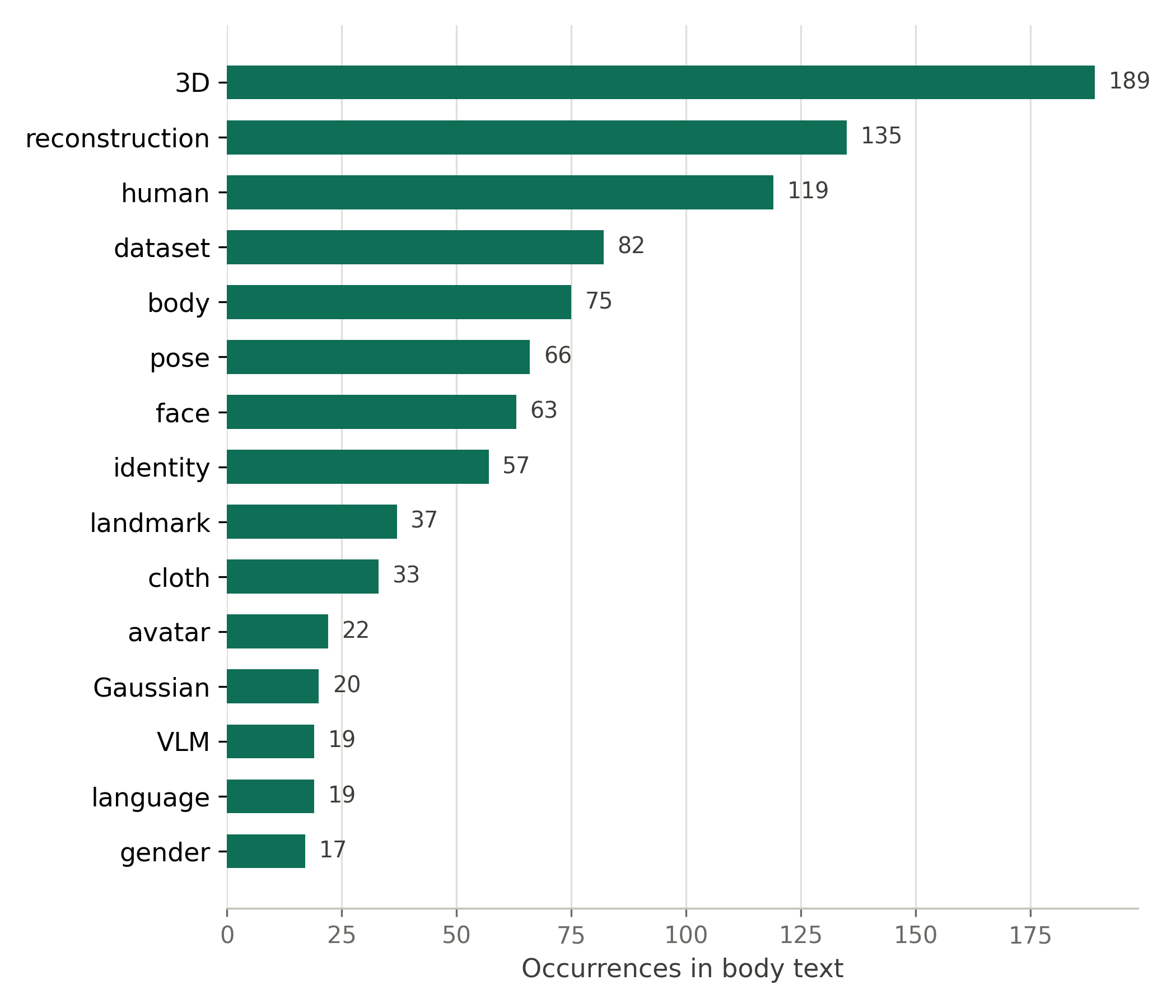}
        \caption{Frequency of key terms.}
    \end{subfigure}
    \hfill
    \begin{subfigure}[b]{0.48\linewidth}
        \includegraphics[width=\linewidth]{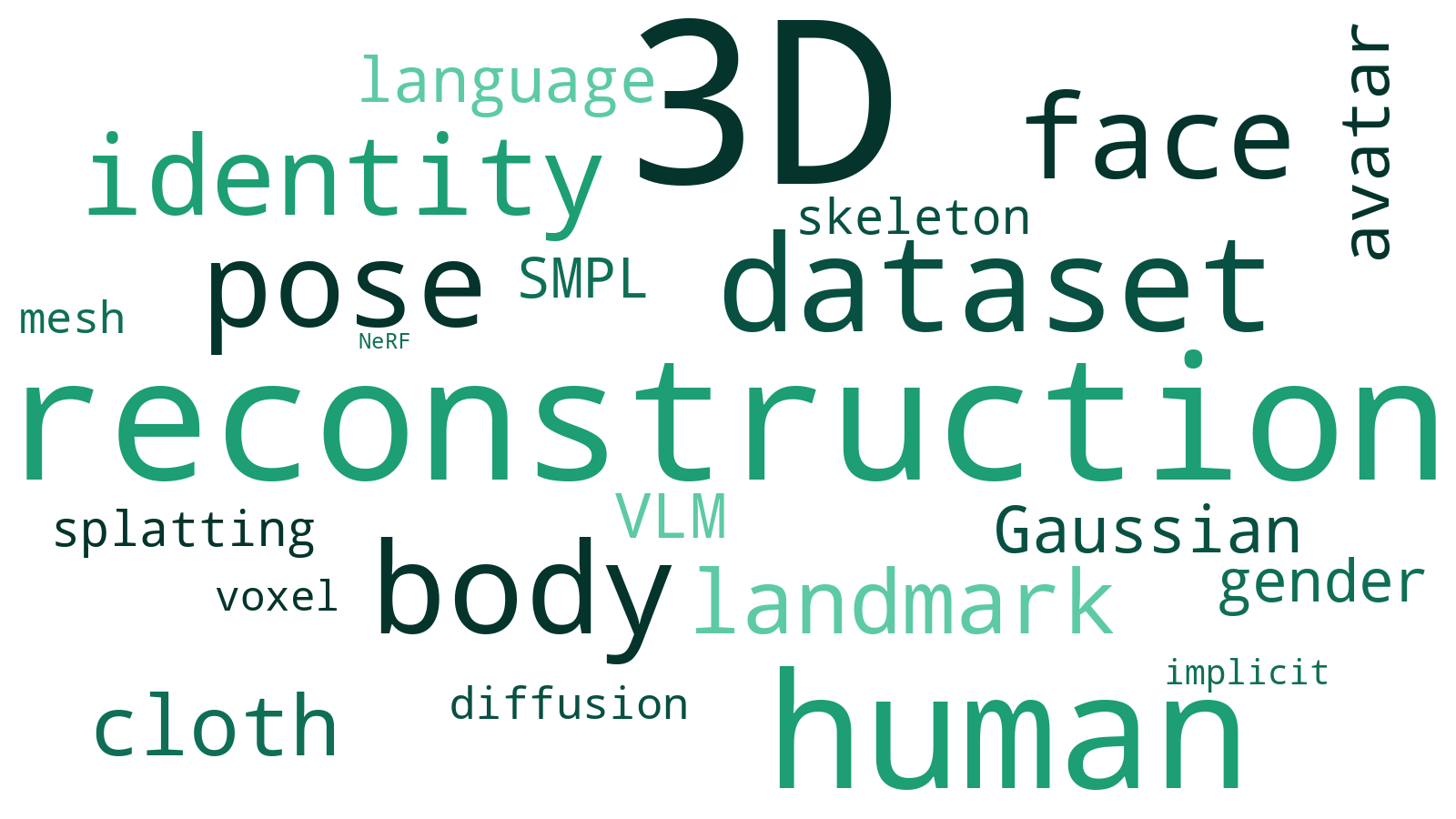}
        \caption{Word cloud, sized by mentions.}
    \end{subfigure}
    \caption{Terminology emphasis across the literature.}
    \label{fig:terminology}
\end{figure*}

We have followed an organized methodology for performing the review. This survey follows a four-stage process: online search, filtration, inclusion or exclusion, and categorization. Fig.~\ref{fig:prisma_funnel} shows this four-stage filteration process included in this review.

\subsection{Search Strategy}
We started our search with different platforms including IEEE Xplore, the CVF, the ACM Library, Springer, NeurIPS Proceedings, and arXiv. We first selected those papers that were  papers published between January 2015 and February 2026.\\
We used the following queries for each of the platforms separately, 
\begin{enumerate}
    \item Q1: ("3D human reconstruction" OR "clothed human avatar" OR "human avatar generation").
    \item Q2: ("identity-aware" OR "identity-preserving") AND ("3D face" OR "3D human")
\item Q3: ("cloth-aware" OR "garment reconstruction" OR "clothed human") AND ("3D Model" OR "avatar")
\item Q4: ("SMPL" OR "parametric human model" OR "human mesh recovery")
\item Q5: ("3D Gaussian Splatting" OR "Gaussian avatar") AND ("human" OR "body")
\item Q6: ("vision-language model" OR "VLM" OR "large language model")  AND ("3D human" OR "avatar" OR "human reconstruction")
\item Q7: ("head pose estimation" OR "facial landmark detection" OR "age and gender estimation") AND ("3D reconstruction" OR "3D modeling")
\end{enumerate}

The papers were searched based on the above queries that returned an initial pool of 112 papers. These papers are further passed through multiple refining steps to reach the final set.

\subsection{Filtering Funnel}
The initial pool of 112 papers is reduced in the following stages. 

\begin{enumerate}
    \item Duplicate removal: this stage removed cross-database and preprint papers, reducing the paper count to 102.
    \item We defined the inclusion and exclusion criteria (defined in the next section). Titles and abstracts were filtered on the basis of the inclusion and exclusion criteria, leaving the count at 97.
    \item Full-text eligibility assessment and quality check based on inclusion and exclusion is ensured that the included papers were about the relevant topic and the quality was good enough to be added to the literature. This step reduced the count to 92.	
\end{enumerate}

\subsection{Inclusion Criteria}
These criteria further filtered the list. A record is retained only if it satisfies the following points.

\begin{enumerate}
    \item Published as a peer-reviewed conference or journal paper, or as an arXiv preprint meeting the quality.
    \item Addresses the target topic in any of the subdomains, including high-level face or body feature for 3D reconstruction, 3D face or body reconstruction, identity-aware or cloth-aware human reconstruction, a benchmark dataset for 3D reconstruction, vision-language-based 3D reconstruction.
    \item Should have a standard qualitative evaluation against a benchmark in the domain.
    \item Should be published between January 2015 and February 2026.
Text should be in English.
\end{enumerate}

\subsection{Exclusion Criteria}
A paper is discarded if it meets any of the exclusion criteria.

\begin{enumerate}
    \item If it addresses generic 3D reconstruction other than human (scene, object or other 3D reconstruction).
    \item Works on body detection or face detection with no work on reconstruction of the human and avatar generation.
    \item In case of duplicate publication, only the peer-reviewed version is retained.
    \item Is it a workshop paper, abstract, or a technical report with no solid methodology presented?
    \item If the paper is from arXiv preprints, then it should be from the last two years. 
\item Full text is not available in English.
\end{enumerate}
\subsection{Categorization}

All the papers that survived the filtering process are categorized into the following six categories: face or body feature extraction,  identity-aware 3D reconstruction, cloth-aware 3D modeling, generative and splatting-based modeling, VLM-based 3D reconstruction, and benchmark datasets for 3D reconstruction. 
\subsection{Limitations of the Search Process} 
We acknowledge multiple limitations to this systematic review in this fast-paced field. These limitations include: restricting to the English language may exclude relevant work published in other languages. Selecting only the last two years of arXiv preprints gives the advantage of selecting top papers, with the assurance that older papers that have not gone through peer review should be discarded. However, this can also exclude important papers.

\section{Literature}
\begin{figure*}
    \centering
    \includegraphics[trim={0cm 0cm 0cm 0cm},width=0.7\linewidth,height=0.7\linewidth,clip]{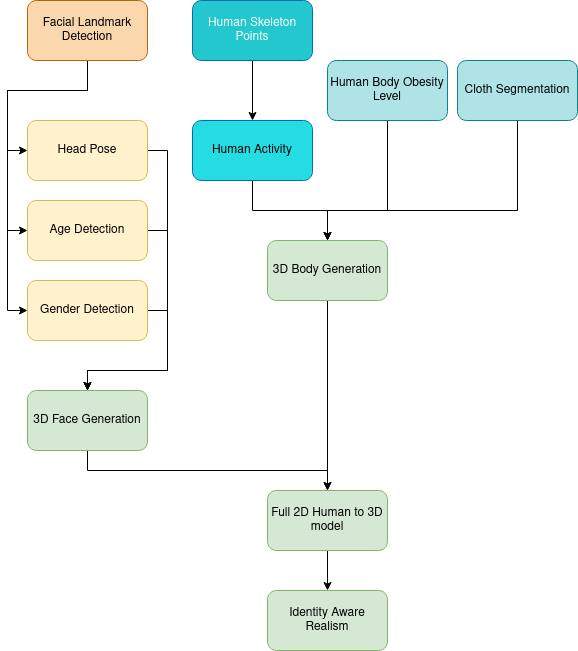}
    \caption{The overall architecture of a general 2d human to 3D model conversion}
    \label{background-architecture}
    \end{figure*}

This section comprehensively reviews related works for all the prerequisites to build identity and cloth-aware 3D human models. The review also revisits several 3D human reconstruction studies concentrating on clothes and identity preservation. \\
Figure \ref{background-architecture} indicates the overall prerequisites involved in identity and cloth aware 3D reconstruction task. The usual breakdown of the cloth-aware and identity-aware 3D  modeling system is as follows,
\begin{enumerate}
      \item{Extract high-level human features}
      \begin{itemize}
            \item{Face Features: Face landmarks (fiducial points), head pose, gender and age.}
            \item{Body Level Features: Skeleton Aware visual body pose and weight.}
       \end{itemize}
       \item{2D to 3D model reconstruction}
       \begin{itemize}
            \item{Simple 3D Modelling}
            \item{Identity Aware 3D Modelling}
            \item {Cloth Aware 3D Modelling}
       \end{itemize}
\end{enumerate}

\subsection {Face level features detection}
Reconstructing an identity- and clothing-consistent 3D human model directly takes advantage of high-level face priors, including age, gender, and landmarks. These parameters control the precise geometric structure in the overall 3D reconstruction task. \\
This section provides a brief review of the role and usage of these high-level features in 3D human reconstruction rather than simple face analysis tasks.\\
Age and gender estimation from a single face image has been proposed using different appearance-based, geometric, and hybrid methods. \\
Appearance-based methods work by extracting color or texture from the face region. These face visual features are then passed through a classifier.  Moghaddam and Yang \cite{10} extracted raw pixel intensities and passed them through a support vector machine to determine gender.  Khan et al. \cite{16} integrated local binary pattern with histogram-of-gradient features for gender and ace/body attribute recognition, respectively. In \cite{16}, they combined body-level features as face-only cues were unreliable for distant subjects. Riera et al. \cite{17} and Dibeklioğlu et al. \cite{18} integrated regional appearance features and dynamic facial deformation cues for age estimation. Similarly, Bukar et al. \cite{19} used appearance and shape features for jointly estimating age and gender. \\
A second type of method employed the geometric position of facial landmarks. Toews and Arbel \cite{14} and Eidinger et al. \cite{15} used landmark-based geometric and hybrid features for high-level attribute extraction under occlusion across different viewpoints. Rizwan et al. [22] built an Active Shape Model from landmarks for age classification using CNN classifier. Wu et al. [24] estimated age using landmark-based face shape on the Grassmann manifold. In a 3D reconstruction system, these high-level features can be used to control and refine the template modification. For example, selecting an age- and gender-based template before mesh and pose fitting can help maintain overall identity.\\
Head pose in terms of yaw, pitch, and roll can control the direct position and structure of the head. These observations are required to align 2D facial representation to a 3D face and head before applying shape- and texture-based deformation. Incorrect pose estimates produce inaccurate and misaligned reconstructions. \\
Landmark-dependent systems first detect facial landmarks and then solve a Perspective-n-Point problem to find pose. Warner et al. [25] extracted pose using facial landmarks and a support vector machine on a synthetic dataset. Baltrušaitis et al. [26] employed Conditional Local Neural Field to regress landmarks and then integrated them with  PnP-based pose recovery. \\
There are a few approaches that work without using landmarks, called landmark-independent approaches. These approaches directly regress pose from image appearance. 6DRepNet [27] regresses a full 6D rotation matrix using a geodesic loss. Ruiz et al. [31] and Hsu et al.'s QuatNet [32] used CNNs directly on images to regress Euler- or quaternion-based pose. WHENet [33] and FSA-Net [34] detected pose using EfficientNet backbones for wide-angle and single-image, respectively.\\
There are reconstruction systems that employ landmarks and pose in a single network in their 3D reconstruction pipelines. MNN [35] and RetinaFace [36] employed pose and landmarks as priors in their reconstruction pipelines, as this removes the separate alignment stage before 3D fitting.
 In practice, head pose estimates can initialize camera-to-face alignment, supporting the non-rigid shape and texture fitting used by identity-aware reconstruction systems. Similarly, pose robustness is a key differentiator between reconstruction systems that can handle in-the-wild capture versus those limited to frontal and controlled environments.

\begin{table*}[]
\begin{tabular}{|l|l|l|l|l|}
\hline
\textbf{Authors}                                                        & \textbf{Type}                                                      & \textbf{Method}                                                                                                                                      & \textbf{Dataset}                                                            & \textbf{Performance}                                                          \\ \hline
\begin{tabular}[c]{@{}l@{}}Moghaddam et al.\\ \cite{10}\end{tabular}     & Gender Classification                                              & Support Vector Machine                                                                                                                               & FERET                                                                       & \begin{tabular}[c]{@{}l@{}}3.4\%\\ (Error)\end{tabular}                       \\ \hline
Geetha et al. \cite{11}                                                 & Gender Classification                                              & \begin{tabular}[c]{@{}l@{}}Local Directional Pattern \\ and LBP features to\\ SVM\end{tabular}                                                       & \begin{tabular}[c]{@{}l@{}}Facial ExpressIon\\ Set (FEI)\end{tabular}       & 99\%                                                                          \\ \hline
Khan et al. \cite{16}                                                  & Gender Detection                                                   & LBP feature to SVM                                                                                                                                   & \begin{tabular}[c]{@{}l@{}}Human attribute\\ and Head-Shoulder\end{tabular} & \begin{tabular}[c]{@{}l@{}}84.8\%\\ 92.0\%\end{tabular}                       \\ \hline
Riera et al. \cite{17}                                                  & Age Group Detection                                                & \begin{tabular}[c]{@{}l@{}}Histogram of Oriented Gradient\\ features to SVM\end{tabular}                                                             & FG-Net aging                                                                & 73.9\%                                                                        \\ \hline
\begin{tabular}[c]{@{}l@{}}Dibeklio lu et al.\\ \cite{18}\end{tabular} & Age Group Detection                                                & \begin{tabular}[c]{@{}l@{}}Facial Dynamic features for \\ different face deformation and \\ appearance features are passed\\ to SVM\end{tabular}     & \begin{tabular}[c]{@{}l@{}}UvA-NEMO Smile\\ UvA-NEMO Disgust\end{tabular}   & \begin{tabular}[c]{@{}l@{}}75.73\%\\ 76.06\%\end{tabular}                     \\ \hline
Bukar et al. \cite{19}                                                  & \begin{tabular}[c]{@{}l@{}}Age and Gender\\ Detection\end{tabular} & \begin{tabular}[c]{@{}l@{}}Appearance Modal using\\ Face Texture and Shape \\ based on Landmarks\end{tabular}                                        & \begin{tabular}[c]{@{}l@{}}FGNET-AD (Gender)\\ HQFace (Age)\end{tabular}    & \begin{tabular}[c]{@{}l@{}}76.65\% (Gender)\\ 4.88\\ Error (Age)\end{tabular} \\ \hline
\begin{tabular}[c]{@{}l@{}}Amna Rizwan et al.\\ \cite{20}\end{tabular}   & Age Group Detection                                                & \begin{tabular}[c]{@{}l@{}}Face Fiducial Points based angle,\\ carnio-facial and wrinkle features\\ to CNN\end{tabular}                              & \begin{tabular}[c]{@{}l@{}}Groups Datasets\\ OUI Dataset\end{tabular}       & \begin{tabular}[c]{@{}l@{}}91.58\%\\ 94.59\%\end{tabular}                     \\ \hline
Wu et al. \cite{22}                                                     & Age Estimation                                                     & \begin{tabular}[c]{@{}l@{}}Face Landmarks and Singular\\ Vector Decomposition based \\ Grassman features are passed\\ to SVM Regression\end{tabular} & FG-Net                                                                      & 6.89 (Error)                                                                  \\ \hline
\end{tabular}
\caption{\label{demo-table}Face Base High Level Features Comparison}
\label{face-highlevel-comparison}
\end{table*}

\begin{figure*}[t]
\includegraphics[trim={0cm 0cm 0cm 0cm},width=1\linewidth]{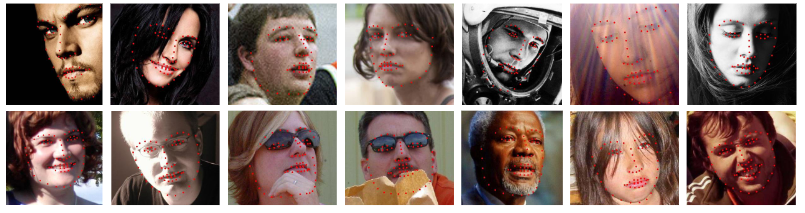}
\caption{Face Landmarks Detection in Different Poses and Occlusion \cite{27}}
\label{landmarks-example}
\end{figure*}

\begin{table*}[tbp]
\hspace{-2em}
\begin{tabular}{|l|l|l|l|l|}
\hline
\textbf{Authors} & \textbf{Type} & \textbf{method} & \textbf{Dataset} & \textbf{Performance} \\ \hline
Warner et. al. \cite{23} & \begin{tabular}[c]{@{}l@{}}Landmarks based \\ head-pose detection\end{tabular} & \begin{tabular}[c]{@{}l@{}}Landmarks are passed \\ to SVM as features set\end{tabular} & \begin{tabular}[c]{@{}l@{}}Synthetic Landmarks\\  based Head-pose (SyLaHP)\end{tabular} & 0.752 AUC \\ \hline
6DRepNet \cite{24} & \begin{tabular}[c]{@{}l@{}}Landmarks independent\\ pose detection\end{tabular} & \begin{tabular}[c]{@{}l@{}}Geodesic loss based \\ regression CNN for \\ head-pose detection\end{tabular} & \begin{tabular}[c]{@{}l@{}}ALFW2000\\ BIWI\end{tabular} & \begin{tabular}[c]{@{}l@{}}3.97 MAE\\ 3.47 MAE\end{tabular} \\ \hline
OpenFace \cite{25} & \begin{tabular}[c]{@{}l@{}}Landmarks based \\ head-pose detection\end{tabular} & \begin{tabular}[c]{@{}l@{}}Conditional Local Neural \\ Field based landmarks \\ regression and PnP based\\  head-pose detection\end{tabular} & \begin{tabular}[c]{@{}l@{}}BIWI\\ BU\end{tabular} & \begin{tabular}[c]{@{}l@{}}2.6 MAE\\ 2.0 MAE\end{tabular} \\ \hline
Multi-Loss ResNet50 \cite{29}& \begin{tabular}[c]{@{}l@{}}Indoor Euler pose angles\\ using a mean squared error\end{tabular} & \begin{tabular}[c]{@{}l@{}}ResNet50 architecture\\ based on multiple mean \\ squared and cross entropy\\ losses\end{tabular} & \begin{tabular}[c]{@{}l@{}}ALFW2000\\ BIWI\end{tabular} & \begin{tabular}[c]{@{}l@{}}6.15 MAE\\ 4.89 MAE\end{tabular} \\ \hline
QuatNet \cite{30} & \begin{tabular}[c]{@{}l@{}}Pose angles using\\ directly from CNN\end{tabular} & \begin{tabular}[c]{@{}l@{}}CNN followed by Regression\\ Net and Ranking Net\end{tabular} & \begin{tabular}[c]{@{}l@{}}ALFW200\\ BIWI\end{tabular} & \begin{tabular}[c]{@{}l@{}}4.50 MAE\\ 4.14 MAE\end{tabular} \\ \hline
WHENet \cite{31} & \begin{tabular}[c]{@{}l@{}}Pose angles using\\ directly from CNN\end{tabular} & \begin{tabular}[c]{@{}l@{}}EfficientNet as a backbone \\ along with three fully \\ connected branches for\\ each respective pose angle\end{tabular} & \begin{tabular}[c]{@{}l@{}}ALFW200\\ BIWI\end{tabular} & \begin{tabular}[c]{@{}l@{}}4.83 MAE\\ 3.48 MAE\end{tabular} \\ \hline
MNN \cite{34} & \begin{tabular}[c]{@{}l@{}}Landmarks detection and\\  head pose detection\end{tabular} & \begin{tabular}[c]{@{}l@{}}Encoder-decoder based \\ architecture with residual\\ blocks for multiple tasks of\\ head pose and landmarks\\ detection\end{tabular} & \begin{tabular}[c]{@{}l@{}}ALFW200\\ BIWI\end{tabular} & \begin{tabular}[c]{@{}l@{}}3.83 MAE\\ 3.66 MAE\end{tabular} \\ \hline
RetinaFace \cite{39}& \begin{tabular}[c]{@{}l@{}}Landmarks detection using\\ CNN and headpose based\\ on detected landmarks\end{tabular} & \begin{tabular}[c]{@{}l@{}}Feature Pyramid Network\\ based 1K landmarks mesh\\ detection and estimation of\\ rotation matrix using least\\ squares solution between\\ regressed landmarks and\\ frontal template landmarks\end{tabular} & AFLW2000 & 3.48 MAE \\ \hline
\end{tabular}
\caption{\label{pose-table}Landmarks and Head Pose Detection Comparison}
\end{table*}

\subsection{Body Level Feature Detection}

Based on figure \ref{background-architecture}, we performed literature and reviewed different body-based features including body skeleton and body weight.\\

\subsubsection{Skeleton Aware Body Pose Detection}
Extracting 3D human body pose based on skeleton is a major pre-requisite to the 3D human body reconstruction. To depict human pose and action in a 3D model, it is essential to first capture movement and placement of all possible joints. A recent effort has been made in this direction by the YOLO family, YOLO-Pose \cite{36} proposed two-dimensional multi-person body pose detection based on joint detection using YOLO object detection paradigm \cite{37}. Their trained model performs human detection and joint prediction in a single network forward pass. The system estimates 17 body keypoints in the 2D domain along with associated confidence for each. YOLO-Pose used CSP-darknet \cite{38} as the backbone network to extract features, PANet \cite{39} for aggregating the backbone features.\\
BlazePose \cite{40} is another recent attempt towards body pose detection using a light convolution neural network. This system is capable of running 30 frames per second and produces 33 different body keypoints. They employed heatmaps and neural regression to predict body coordinates. In the training phase, heatmaps and offset maps are used exclusively to fine-tune the embeddings for the encoder network. This system utilized skip connections for propagating local features deep in the network and it also employs a unique mechanism of stop gradient where features are passed in forward direction.\\\\
Some of the researchers focused on multi-person 2D pose estimation (MPPE). Hidalgo G et al. \cite{41} presented a multi-person skeleton detection approach based on a single network. They pioneered the idea of single network based skeleton detection. Their proposed network is capable of detecting key points of face, hand, body and feet in a single network. This system employed the Part Affinity Field (PAF) from OpenPose to detect keypoints.\\
Similarly, another research group also worked on multi-human 2D skeleton detection \cite{42} using a single model for the whole body including face, hand, body and feet points. This group also proposed the first dataset for whole body keypoints and named it as COCO WholeBody. They utilized ZoomNet for keypoints detection of different parts.  Another similar attempt \cite{43} is made in HPRNet where hierarchical point regression is applied for detecting full body keypoints in single phase. They utilized the COCO-Wholebody dataset and improved the overall baseline network. \\

\subsubsection{Body Weight Analysis}

Body weight or volume is taken into account by a 3D reconstruction system in order to better reflect the body contour in 3D reconstructed models, leading to a more accurate and customized portrayal. In literature there has been a reasonable amount of work for detection of human weight based on body \cite{44, 45} and face \cite{46,47}.\\
Jiang et al. \cite{44} used body dimensions for detecting the body weights and solved two tasks of BMI and weight classification. They also proposed a self-collected dataset for body weight and BMI, containing 5900 images. This system utilized body contour along with skeleton points as a key feature set for detecting body weight. This system also employed different ratios including waist-height, waist-hip and abdominal sagittal diameter. All these features are passed from Support Vector Regression for finding BMI. Similarly, In \cite{45} anthropometric features are extracted using a neural network regressor including arm length, waist circumference, leg length and leg circumference. These features are further used to deduce weight of the subject.\\ 
Face based obesity detection systems require a face based image in which face is fully detectable and these systems detect face landmarks and try to find the obesity and BMI using face shape using landmarks \cite{46}. Other systems detect weight using a neural network directly by inputting the cropped face into the residual network \cite{47}. Face-based obesity detection systems are good options for face applications but for full body applications face based obesity detection may fail on distant subjects. \\

\begin{figure*}[!t]
    \centering
    \includegraphics[width=\textwidth]{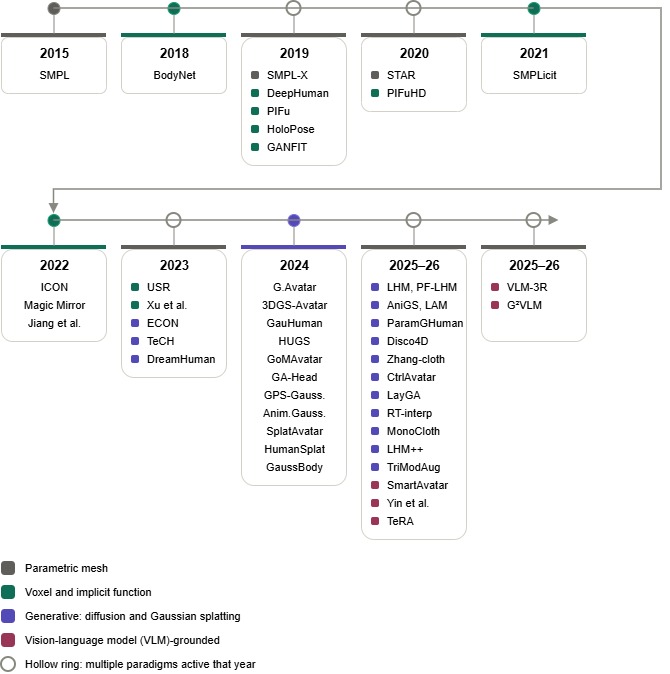}
    \caption{Chronological evolution  of 3D human reconstruction Systems. The timeline is from parametric mesh and implicit-function methods to generative Gaussian Splatting and, since 2025, vision-language model-grounded reconstruction.}
    \label{fig:3d_reconstruction_progression}
\end{figure*}

\subsection{Human Features based 3D modeling}
After reviewing all the pre-requisite supporting features, we reviewed 3D reconstruction systems. During the last decade, many different networks have been proposed to project 2D pose to 3D model \cite{1a,2a,3a}.  Wang et al. \cite{3a} worked on learning the joint distribution and 3D pose estimation. Their major contribution was to develop a pose estimation system that captures the joint information and maps it to 3D space. They proposed a single-stage (DAS) model for finding the position and mapping it to 3D pose in a single pass. Similarly, Ruixu Liu et al. \cite{1a} worked on 3D human pose estimation using Attention-Based Neural Network. Their major contributions involve joint detection followed by 3D pose projection using an attention mechanism. There are other researchers who worked on 3D human pose estimation and modeling \cite{4a,5a}. \\
Some researchers worked on the modeling of the face based on features including eyes, lips and hairs \cite{6a,7a}.  Alldieck et al. \cite{6a} proposed a natural face avatar modeling system that captures clothes along with refined facial cuts based on face landmarks. Similarly, Zhang et al. \cite{7a} worked on producing face based animation by capturing face features along with emotion. Some of the researchers focused on body appearance by maintaining the cloth depiction \cite{8a,9a}. Table \ref{tab:comparison} presents the summary of different identity and cloth-aware 3D reconstruction systems. \\

\begin{figure*}[t]
\centering
\includegraphics[trim={0cm 0cm 0cm 0cm},width=0.8\linewidth,height=0.5\linewidth]{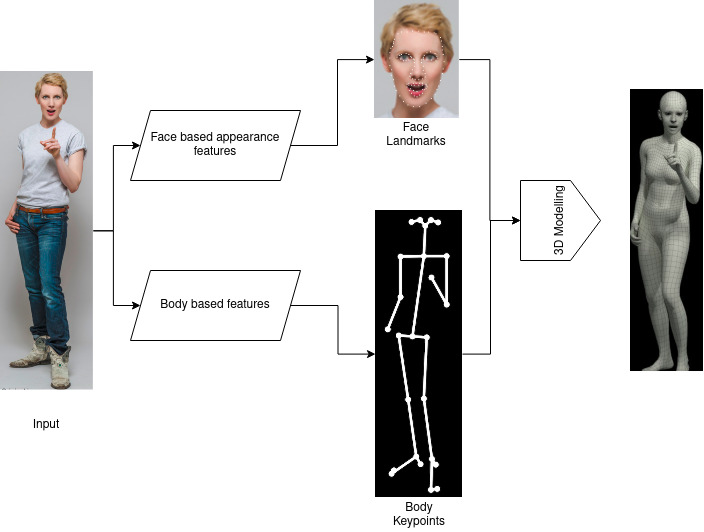}
\caption{Identity Aware 3D Reconstruction of human based on face and body features}
\label{identity-aware-architecture}
\end{figure*}
\subsubsection{Identity Aware 3D modeling}

Preserving the identity in 3D reconstruction is important to provide a customized perspective. The typical structure of identity-aware 3D reconstruction system is shown in figure \ref{identity-aware-architecture}. Different research groups worked on full body 3D reconstruction by preserving facial structure \cite{48, 49}. Preserving identity from distant subjects is a challenging task, some of the researchers \cite{50, 51, 52*} worked only on 3D modeling of face. They preserved the face identity and facial features depicted in the converted model.\\
\textbf{Mesh}: Earlier attempts for 3D reconstruction are Skinned Multi-Person Linear Model (SMPL) \cite{54} based but they lack in capturing facial details and garments details. HoloPose \cite{48} proposed a part-based 3D model reconstruction. They improved the existing  SMPL mesh \cite{52} to handle occlusion and rotation by proposing part-based features extraction around human joints. As the keypoints varies by changing the position and the local features remained invariant to the changing angle and occlusion. Fully convolutional neural network detects the joints location followed by a polling layer to select the features from locality of the joints that provides local information of the source subject. Finally, for each joint an angle is drived to reconstruct the human body with full angles.

\textbf{Voxels}: Voxels are primary units in 3D space for 3D models similar to tiny cubes. In computer graphics voxels grid are widely used instead of meshes. Several research groups worked on voxel-based representation of human body. BodyNet \cite{73} and Gilbert et al. \cite{74} proposed creating volumetric model from visual images. BodyNet employed multi-view re-projection loss and volumetric 3D loss function to train end-to-end neural network for body pose, body parts and 3D reconstruction. Gilbert et al. utilized encoder-decoder network for constructing 3D volumetric representation of images in different view angles. \\

Similarly, DeepHuman \cite{49} also proposed a similar approach to 3D human reconstruction by taking SMPL as base. They extracted semantic volume for parts to construct 3D human. Their proposed network consists of three modules, image encoder, volume-to-volume translation and refinement network. These three modules enabled the system to depict the volume of the source subject in the target model.\\

\textbf{Implicit functions}: For near field images Xu et al. \cite{51} proposed a 3D human head reconstruction system that preserve the identity by employing neural implicit function. This system decomposes the geometric features into identity-independent and dependent modules. Identity independent module is a smooth surface that acts as a template of the face including outer cuts of the face. Identity dependent components compute the displacement fields to capture facial features including wrinkles, fine cuts and small face features. Similarly, GANFIT \cite{50} utilized a Generative Adversarial Network for 3D Morphable Model fitting to the task of 3D face reconstruction. They preserve identity by utilizing face recognition loss function to keep the identity of source in 3D model. \\
In \cite{52*}, shape and identity aware regularization loss is employed to better train shape features to preserve identity in the reconstruction task. This loss contains of two main components: identification loss, which accounts for interclass and intraclass variability, and the overall distribution of 3D morphable model parameters. This system employed a deep neural network trained on face identity dataset to comprehend the correlation between shape attributes and the facial geometry.\\
\BeginOneColumn
\begingroup
\setlength{\tabcolsep}{3pt}
\small
\begin{longtable}{|p{1.9cm}|p{1.8cm}|p{1.8cm}|p{3.2cm}|p{2.3cm}|p{3.0cm}|}
\caption{Comparison of surveyed 3D human reconstruction methods across representation
paradigms. Note that performance metrics are heterogeneous across rows (e.g., PA-MPJPE,
IOU, Chamfer distance, PSNR are not directly comparable) since each subfield has adopted
different evaluation protocols; values should be read only within their representation
category, not across the table.} \label{tab:comparison} \\
\hline
\textbf{Authors} & \textbf{Representation} & \textbf{Type} & \textbf{Method} & \textbf{Dataset} & \textbf{Performance} \\
\hline
\endfirsthead
\multicolumn{6}{l}{\textit{Table \thetable\ continued from previous page}} \\
\hline
\textbf{Authors} & \textbf{Representation} & \textbf{Type} & \textbf{Method} & \textbf{Dataset} & \textbf{Performance} \\
\hline
\endhead
\hline
\multicolumn{6}{r}{\textit{continued on next page}} \\
\endfoot
\hline
\endlastfoot
 
\hline
\multicolumn{6}{|l|}{\textbf{Identity-Aware 3D Modeling}} \\
\hline
HoloPose~\cite{48} & Parametric mesh & Identity-aware & Part-based reconstruction via joint locations & Human3.6M & 46.52 (PA-MPJPE) \\
\hline
GANFIT~\cite{50} & Parametric mesh & Identity-aware face & GAN + face-recognition loss & MICC Florence 3D Faces & 0.95 mean, 0.107 std (point-to-plane) \\
\hline
Xu et al.~\cite{51} & Parametric mesh & Identity-aware face & Identity-independent template + displacement field & Portrait Relighting Dataset & 0.930 / 0.785 (Chamfer, 10-view) \\
\hline
Jiang et al.~\cite{52*} & Parametric mesh & Identity-aware face & Convolution-based encoding & LFW, CFP-FP & 94.47 / 80.78 (Euclidean) \\
\hline
DeepHuman~\cite{49} & Voxel & Identity + volume-aware & Image encoder + volume-to-volume + refinement & THuman & 45.7\% (IOU) \\
\hline
 
\hline
\multicolumn{6}{|l|}{\textbf{Cloth-Aware 3D Modeling}} \\
\hline
3D Magic Mirror~\cite{53} & Parametric mesh & Cloth-aware & Four encoders: shape, pose, illumination, texture & ATR, Market-HQ & 81.1 / 83.4 (IOU) \\
\hline
USR~\cite{54} & Implicit function & Cloth-aware & Encoder--decoder, decoupled cloth/body & Diverse Dressed People & 0.00028 (Chamfer distance) \\
\hline
SMPLicit~\cite{55} & Parametric mesh & Cloth-aware & Separate encoders for cut/style/body & BCNet & - \\
\hline
 
\hline
\multicolumn{6}{|l|}{\textbf{Generative and Splatting-Based 3D Human Modeling}} \\
\hline
GaussianAvatar~\cite{gaussianavatar} & 3D Gaussian Splatting & Animatable avatar & Gaussians mapped to SMPL skeleton parameters & NeuMan, People-Snapshot, DynVideo & PSNR 28.90, SSIM 0.969, LPIPS 0.024 (NeuMan) \\
\hline
3DGS-Avatar~\cite{3dgsavatar} & 3D Gaussian Splatting & Animatable avatar & Gaussians mapped to SMPL skeleton parameters & ZJU-MoCap & 30 (PSNR), 0.9703 (SSIM), 29.58 (LPIPS) \\
\hline
GauHuman~\cite{gauhuman} & 3D Gaussian Splatting & Reduced-view reconstruction & Monocular video, fewer input views & ZJU-MoCap, MonoCap & PSNR 31.34 / 33.45; 189 FPS \\
\hline
Animatable Gaussians~\cite{animatable} & 3D Gaussian Splatting & Reduced-view reconstruction & Pose-dependent Gaussian maps &  THuman4.0, ActorsHQ & 8.0714 PSNR, 0.9739 SSIM, 0.0515 LPIPS\\
\hline
HUGS~\cite{hugs} & 3D Gaussian Splatting & Human + scene & Joint human/background reconstruction & NeuMan, ZJU-MoCap &  30.54 PSNR, 0.97 SSIM, 0.03 LPIPS\\
\hline
GPS-Gaussian~\cite{gpsgaussian} & 3D Gaussian Splatting & Feed-forward & Single forward-pass Gaussian prediction & Twindom, THuman2.0 & 25.05 PSNR, 0.886 SSIM, 0.121 LPIPS \\
\hline
HumanSplat~\cite{humansplat} & 3D Gaussian Splatting & Feed-forward & Single forward-pass Gaussian prediction & Twindom, THuman2.0 & 24.293 PSNR, 0.915 SSIM, 0.040 LPIPS \\
\hline
GaussianBody~\cite{gaussianbody} & 3D Gaussian Splatting & Clothed-human & 3DGS for clothed-human reconstruction & PeopleSnapshot, iPER & 35.66 PSNR, 0.9753 SSIM, 0.021 LPIPS \\
\hline
LHM~\cite{lhm} & 3D Gaussian Splatting & Feed-forward, animatable & Multimodal body--head transformer & 200 in-the-wild videos & PSNR 22.00, SSIM 0.930, LPIPS 0.040 \\
\hline
PF-LHM~\cite{pflhm} & 3D Gaussian Splatting & Pose-free & Pose-independent extension of LHM &  in-the-wild fashion video dataset & 22.363 PSNR, 0.887 PSNR, 0.057 LPIPS \\
\hline
Disco4D~\cite{disco4d} & 3D Gaussian Splatting & Layered avatar & Decoupled body + garment Gaussians & Synbody, CloSe & 15.691 PSNR, 0.848 SSIM, 0.185 LPIPS \\
\hline
Zhang et al.~\cite{zhangcloth} & 3D Gaussian Splatting & Layered avatar & Decoupled body + garment (single-model) & THuman2.0 & 12.50 FID, 9.39 KID\\
\hline
MonoCloth~\cite{monocloth} & 3D Gaussian Splatting & Cloth-decoupled & Full-body cloth decoupling + reanimation & X-Humans, NeuMan & 33.53 PSNR, 0.986 SSIM, 1.20 LPIPS\\
\hline
TeCH~\cite{tech} & Diffusion prior & Text/image-conditioned & Diffusion prior for geometry + texture & CAPE, THuman2.0 &   25.2107 PSNR, 0.9363 SSIM, 0.0835 LPIPS\\
\hline
ICON~\cite{icon} & Implicit function & Occlusion-robust & Normal-map-guided implicit reconstruction & THuman2.0 / CAPE & 1.142 (Chamfer) \\
\hline
ECON~\cite{econ} & Implicit + explicit & Occlusion-robust & Normal-map surfaces + SMPL-X inpainting & THuman2.0 / CAPE, Renderpeople & $-$38.2\% Chamfer, $-$57.6\% normal error \\
\hline
DreamHuman~\cite{dreamhuman} & Diffusion (SDS) & Text-to-avatar & Score-distillation sampling from text & 160 text prompts (CLIP eval); CMU MoCap (pose sampling) &  R-Prec. 0.838, Top-3 0.931, Top-5 0.956 \\
\hline
 
\hline
\multicolumn{6}{|l|}{\textbf{Vision-Language Model (VLM)-Based Human Body Reconstruction}} \\
\hline
xu et al.~\cite{xu} & VLM-grounded & Text-guided mesh recovery & Per-body-part text descriptions as constraints &  Human3.6M, 3DPW &  3DPW MPJPE 69.3, 3DPW PA-MPJPE 43.9,  Human3.6M MPJPE 47.7, Human3.6M PA-MPJPE 32.4 \\
\hline
SmartAvatar~\cite{smartavatar} & VLM-grounded & Agentic refinement & VLM verify-and-refine loop & No fixed benchmark & 0.65 ArcFace ID Similarity, 0.903 CLIP\textsubscript{image}  \\
\hline
MotionGPT~\cite{motiongpt} & VLM-grounded (motion) & Motion--language & Motion tokenization for language generation & HumanML3D, KIT & 0.492 R TOP, 0.232 FID, 9.528 DIV \\
\hline
MotionLLM~\cite{motionllm} & VLM-grounded (motion) & Motion understanding & Motion/video understanding via VLM & MoVid & 49.0 Acc, 2.97 Score \\
\hline
3D-LLM~\cite{3dllm} & VLM-grounded & 3D-language foundation & 3D features projected into LLM input space & ScanQA, SQA3D, 3DMV-VQA & ScanQA BLEU-1 37.5 ($+$9\%), CIDEr 74.5 ($+$10\%) \\
\hline
VLM-3R~\cite{vlm3r} & VLM-grounded & Instruction-tuned 3D & Instruction-aligned 3D reconstructive tuning & VSI-Bench & 60.90 Accuracy \\
\hline
G\textsuperscript{2}VLM~\cite{g2vlm} & VLM-grounded & Spatial reasoning & VLM for spatial reasoning + reconstruction & ScanNet, Co3Dv2, BlendedMVS, DL3DV, MegaDepth, ArkitScenes, HyperSim, ScanNet++, Habitat, TartanAir &  SPAR-Bench 54.87 \\
\hline
TeRA~\cite{tera} & VLM-grounded & Text-to-realistic avatar & Realism-focused text-to-3D refinement & HuGe100K & 30.17 CLIP Score, 0.82 VQA Score  \\
\hline
 
\end{longtable}
\endgroup
\EndOneColumn
\subsection{Generative and Splatting-Based 3D Human Modeling}
Another direction for 3D human reconstruction is generating meshes through a diffusion process rather than direct regression.\\
3D Gaussian Splatting (3DGS) represents a scene as a set of anisotropic 3D Gaussians captured directly, without costly volumetric sampling. Based on 3D Gaussian Splatting, GaussianAvatar [95] and 3DGS-Avatar [96] mapped a set of Gaussians to SMPL skeleton parameters. They allowed the avatar to change pose in animation in real time. This unique property is not offered by implicit- and voxel-based methods without heavy post-processing. \\
GauHuman [97] and Animatable Gaussians [98] reduced the number of views for 3D reconstruction using monocular video, addressing the limitation of multi-view stream availability. HUGS [99] also worked on the same issue and reconstructed 3D human and background scene from monocular sequence. Enhancing this, GPS-Gaussian [100] and HumanSplat [101] worked on predicting Gaussian parameters with a single forward pass without subject-based optimization. GaussianBody [102] employed 3D Gaussian Splatting (3DGS) for clothed-human reconstruction that PIFu and DeepHuman originally addressed. LHM [103] further advanced this simple feed-forward strategy in 2025. They employed a multi-modal transformer and mapped single image directly to an avatar for animation purpose. Facial identity was preserved by a feature-pyramid encoder. This identity preservation is substantially faster than the identity-preservation loss used in GANFIT [60].  PF-LHM [104] further extended the process to pose-independent reconstruction. Their pose-free reconstruction could reconstruct from in-the-wild images without explicitly requiring body-pose annotations. \\
Disco4D [105] and  Zhang et al. [106] worked on body and garment Gaussian parameters and treated the cloth-aware reconstruction as a native single model recontruction rather than treating a clothing as a separate layer over SMPL based body layer  as in SMPLicit [68]. MonoCloth [107] further worked on cloth-aware reconstruction and achieved full-body cloth decoupling and reanimation from monocular video.\\
A parallel direction abandons regression in favor of generative diffusion. 
Another direction employs diffusion methods for geometry reconstruction and texture generation. TeCH [108] uses a diffusion prior on a single input image to estimate the subject's overall geometry and texture across the whole body, including unseen views like the back, occluded areas, and self-occlusion.  ICON [109] and its successor ECON [110] employed a normal map to achieve robust reconstruction under occlusion with different clothing.  DreamHuman [111] generated an identity- and clothing- consistant 3D model from a text prompt.\\
DreamHuman [111] reconstructed animitable 3D human avatar from a textual prompt. They used score-distillation sampling with a pretrained 2D diffusion model, removing the requirement for providing an input image. \\
Taken together, these discussed methods include real-time rendering through Gaussian Splatting, feed-forward generalization through large reconstruction models such as LHM, and generating unseen views advanced by diffusion priors. These methods represent the active research direction in the field of identity- and cloth-aware 3D human reconstruction.
\subsubsection{Cloth Aware 3D modeling}
Using encoder-decoder networks as a main tool, 3D clothing reconstruction seeks to produce the geometry and texture of clothes \cite{53, 54, 55}. Such systems employ an encoder to convert images into feature sets, which are then reconstructed into a 3D model by a decoder. The decoder performs the role of a rendering engine, frequently using previously acquired information from 3D templates to precisely represent the geometry and texture of clothing. While some researchers have created ways to produce cloth-aware 3D models from a single image \cite{57, 58}, others have concentrated on video-based 3D reconstruction while maintaining cloth appearances details \cite{54, 56}.\\
3D Magic Mirror \cite{53} introduced a 3D cloth aware human reconstruction system from 2D image using four special tasked encoders. Their proposed system encoded features for shape, pose, illumination and texture in a separate dedicated encoder. They employed a differentiable renderer-based decoder that combines the shape, pose, illumination and texture features to render a 3D model. They also produced the foreground mask during the reconstruction task as an extra task.\\
Similarly, USR \cite{54} worked on reconstructing the dressed human images. They utilized multi-view frames of human to learn the geometry of the dressed body. They proposed a layered based system for 3D model and cloth layers separately. This system is unsupervised in nature and consists of three subcomponents 1) Generalized Surface aware Neural Radiance Field (GSNeRF) for the purpose of constructing surface, 2) Semantic and Confidence Guided Separation strategy (SCGC) for modeling 3D garments separately, 3) Edge refinement model to smooth out the edges as a post-processing step.\\
SMPLicit \cite{55} proposed a generative model that is trained to learn body pose, shape and different types of clothes at the same time. Model learns the attributes of upper-body cloths and lower-body cloths separately. To control the cloth cut and style, they utilized a different set of features along with body shape parameters.
\subsection{Vision-Language Model-Based Human Body Reconstruction}
These types of models are very recent and attach language reasoning to the input image using a vision backbone for the purpose of human reconstruction. Rather than treating human reconstruction as a pure geometric task, there is growing interest in using a vision-language model (VLM) to supervise the reconstruction procedure with semantic feedback and to create body pose and action using natural language. Xu et al. [112] reformulated the problem of human reconstruction by generating per-body-part semantic descriptions to guide the optimization process in reconstruction  (for example, describing the bend of an arm or the orientation of the legs and hips). Their pipeline includes a text encoder and a pose Vector Quantized Variational AutoEncoder (VQ-VAE) to map these descriptions to the body pose in a shared latent space. This description can handle occlusion and depth ambiguity in an agentic learning loop rather than using regression-based techniques. \\
This is conceptually similar to the natural-language instructions for action generation in VLA models, but here these instructions are applied to 3D reconstruction refinement rather than controlling the robot. SmartAvatar [113] goes beyond using these instructions as a training-time guide signal and places a VLM at the core of an agentic reconstruction loop. They used a textual prompt and an input image to produce an avatar using an off-the-shelf engine and a VLM model that evaluates the face identity, structural plausibility, and prompt reflection. VLM acts as an agent that iteratively adjusts generation parameters until the generated mesh is identity- and prompt-consistent. This refinement loop reflects the strategy of perceive-reason-act in hierarchical planning agents, applied here to refine the avatar identity and structural accuracy. \\
Another related direction is to capture human motion rather than simple shape reconstruction using language. MotionGPT [115] represented human movement as language representations. They tokenized the pose trajectories to generate captions and answer questions about human motion using a VLM. Another similar architecture, MotionLLM [116] employed these methods to analyze human behavior from motion and video to answer questions. These systems work on motion generation and understanding rather than generating static 3D models. The tokenization process bridges the body representation with LLM-based reasoning. 3D-LLM [117] used this approach in 3D spaces by integrating 3D space features into the LLM's input space to support question answering and grounding in 3D space.\\
VLM-3R [118] extends a VLM with instruction-aligned 3D reconstructive tuning. They tried to answer reasoning questions about the reconstructed scene without requiring extra depth information. G\textsuperscript{2}VLM [119] used VLM for spatial reasoning and 3D reconstruction. \\
 Neither system targets human-body reconstruction specifically, but gives a direction for human reconstruction pipelines that can be useful for reasoning and refined through natural-language interaction rather than using geometric loss. Finally, TeRA [120] improved the text-based 3D reconstruction with a focus on realism and plausibility. They addressed the failure scenarios of CLIP- and diffusion-driven text-to-avatar systems.\\
Collectively, this direction suggests that future identity- and cloth-aware 3D human reconstruction systems will not only reconstruct in a specific domain (mesh, implicit function, or Gaussian) but also integrate the language-reasoning layer to supervise, verify, and ground that reconstruction.\\
Some of the above systems use tokenization to represent the 3D human body and rely on a language model. 
 Yin et al. \cite{xu} use a pose VQ-VAE to represent the human body as discrete tokens. MotionGPT \cite{motiongpt} tokenizes motion, and VQ-HPS uses tokens to construct a 3D body through a learned codebook. However, a learned codebook can collapse, and errors can propagate into reconstruction. EgoTokens \cite{egotokens} tokenizes the 122 SMPL-X parameters to one of 1,000 uniform bins. It uses no learned codebook, which prevents collapse, and each token is directly readable as a parameter value. It also uses a distance-aware loss that penalizes near bins less and far bins more. None of the other surveyed papers use a distance-aware VLM loss for 3D reconstruction. 

\begin{figure*}[t]
    \centering
    \includegraphics[width=0.95\textwidth]{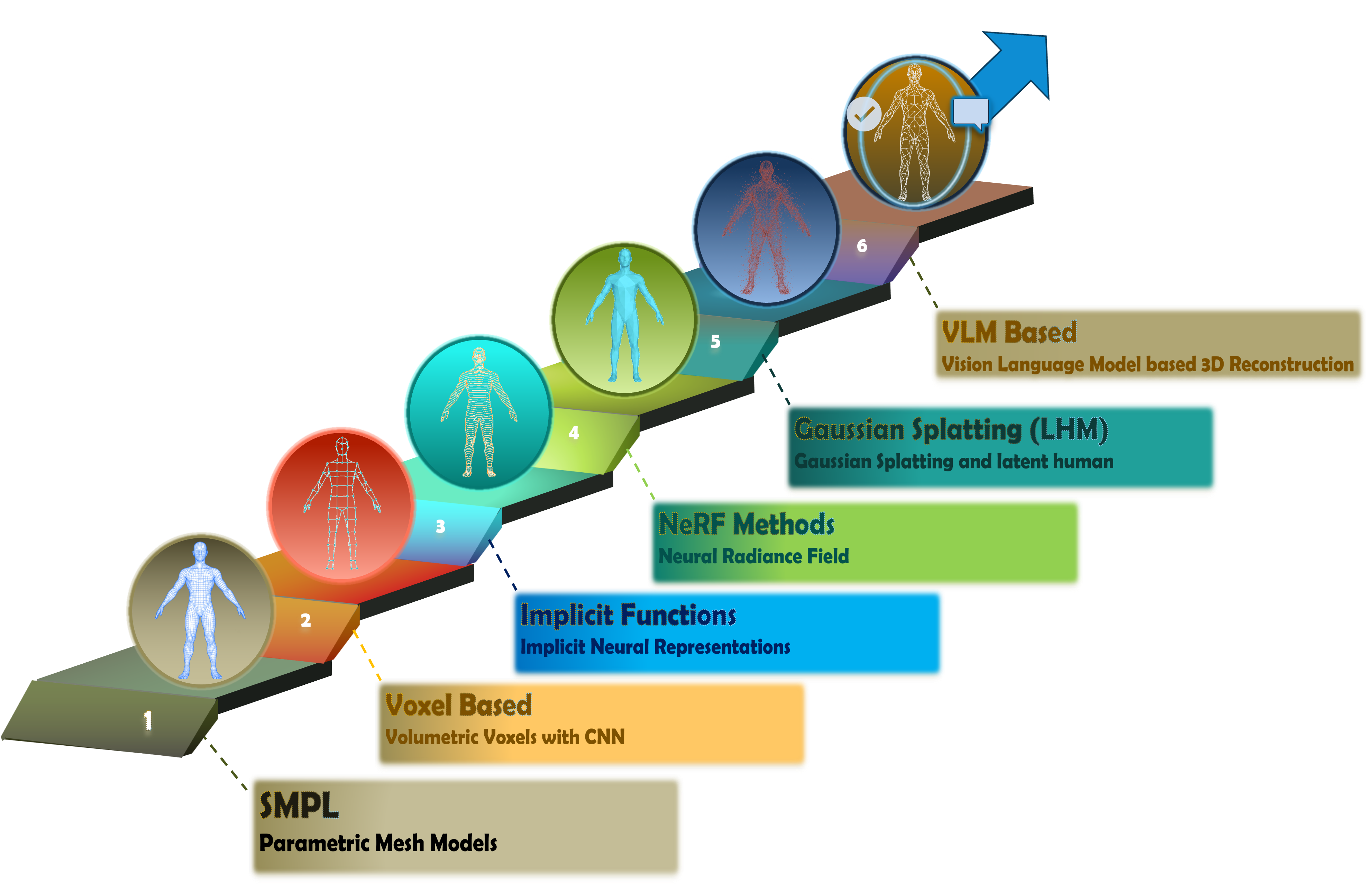}
    \caption{Evolution of 3D human body representations, from parametric mesh to vision-language model-grounded reconstruction (2015–2026)}
    \label{fig:full_progression}
\end{figure*}

\section{3D Modeling Datasets}

\begin{figure}
\begin{subfigure}{.2\textwidth}
  \centering
  \includegraphics[width=.8\linewidth]{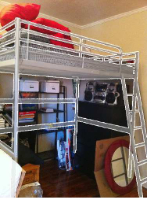}
  \caption{Pix3D}
  \label{fig:sfig1}
\end{subfigure}%
\begin{subfigure}{.3\textwidth}
  \centering
  \includegraphics[width=.6\linewidth]{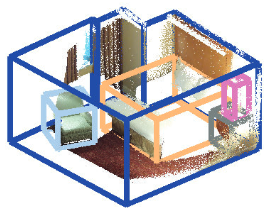}
  \caption{Sun RGB-D}
  \label{fig:sfig2}
\end{subfigure}
\begin{subfigure}{0.2\textwidth}
  \centering
  \includegraphics[width=0.9\linewidth]{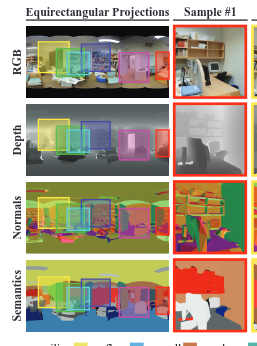}
  \caption{Joint 2D-3D Data}
  \label{fig:sfig2}
\end{subfigure}
\begin{subfigure}{0.2\textwidth}
  \centering
  \includegraphics[width=0.9\linewidth]{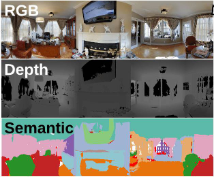}
  \caption{Matterport3D}
  \label{fig:sfig2}
\end{subfigure}

\begin{subfigure}{.23\textwidth}
  \centering
  \includegraphics[width=1\linewidth,height=.6\linewidth]{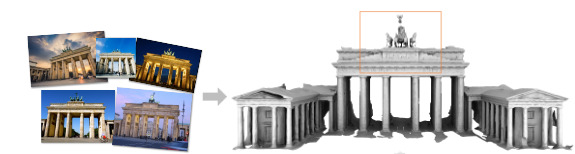}
  \caption{Heritage Recon}
  \label{fig:sfig2}
\end{subfigure}
\begin{subfigure}{.23\textwidth}
  \centering
  \includegraphics[width=1\linewidth,height=.6\linewidth]{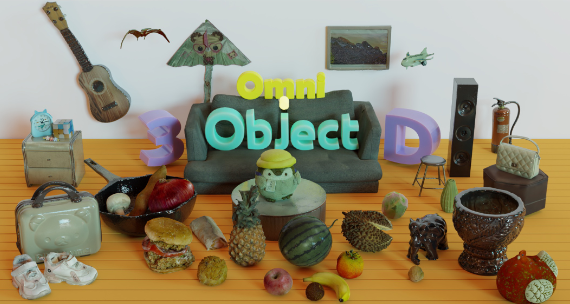}
  \caption{OmniObject3D}
  \label{fig:sfig2}
\end{subfigure}

\caption{3D Object Reconstruction Datasets}
\label{objects-datasets}
\end{figure}

\begin{figure*}
\begin{subfigure}{.35\textwidth}
  \centering
  \includegraphics[width=1\linewidth,height=0.7\linewidth]{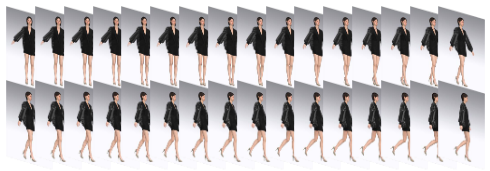}
  \caption{CLOTH4D}
  \label{fig:sfig1}
\end{subfigure}%
\begin{subfigure}{.3\textwidth}
  \centering
  \includegraphics[width=1\linewidth,height=0.7\linewidth]{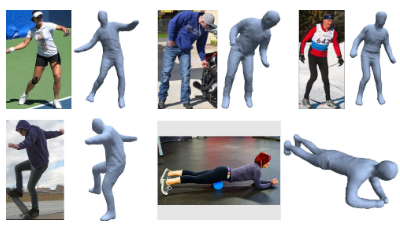}
  \caption{THuman}
  \label{fig:sfig2}
\end{subfigure}
\begin{subfigure}{0.3\textwidth}
  \centering
  \includegraphics[width=1\linewidth,height=0.7\linewidth]{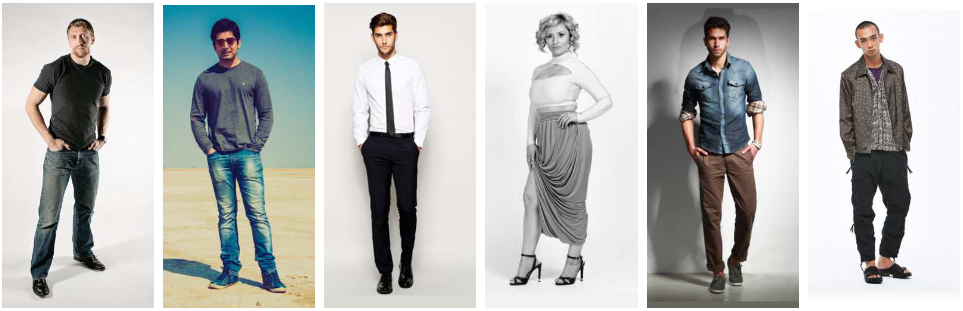}
  \caption{3D Human Body Dataset}
  \label{fig:sfig2}
\end{subfigure}
\begin{subfigure}{0.35\textwidth}
  \centering
  \includegraphics[width=1\linewidth,height=0.7\linewidth]{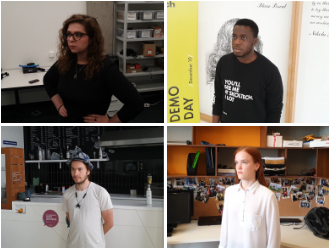}
  \caption{SmartPortraits}
  \label{fig:sfig2}
\end{subfigure}
\begin{subfigure}{.3\textwidth}
  \centering
  \includegraphics[width=1\linewidth,height=.6\linewidth]{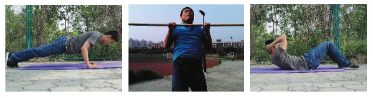}
  \caption{$HSiPu^2$}
  \label{fig:sfig2}
\end{subfigure}
\begin{subfigure}{.35\textwidth}
  \centering
  \includegraphics[width=1\linewidth,height=0.7\linewidth]{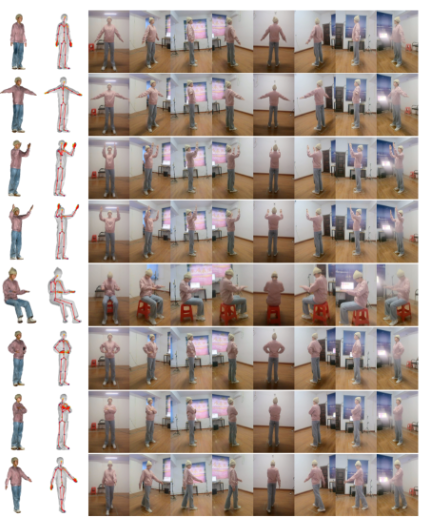}
  \caption{MVP-Human dataset}
  \label{fig:sfig2}
\end{subfigure}

\caption{3D Human Reconstruction Datasets}
\label{human-datasets}
\end{figure*}

Benchmarking datasets have a notable role for validating the overall performance of any reconstruction system. Significant progress has been achieved in 3D reconstruction recently, and multiple 3D datasets for reconstructing human and non-human objects have been produced.(table \ref{dataset-table}). \\
In this section, 3D reconstruction benchmarking datasets have been reviewed based on their type and purpose. Majorly there are two types of the datasets;
\begin{enumerate}
    \item Non-human object based datasets
    \item Human based datasets
\end{enumerate}
Non-human datasets include the reconstruction of indoor and outdoor objects in different scene settings. These datasets do not include humans, some types are shown in figure \ref{objects-datasets}. On the contrary, 3D human datasets include human as an imperative part of the frame. Figure \ref{human-datasets} depicts examples from human based datasets.\\
Objects based non-human datasets \cite{59, 60, 61, 62, 63} provide different 3D models with different styles and scenes. Mostly these datasets are used for evaluating the shape of 3D reconstructed objects and sometimes they are employed for classification tasks. On the contrary, there are a variety of datasets available that focus on human as a primary part. These datasets vary in different types including clothed human, simple human and face based. All these datasets \cite{64, 65, 66, 67, 68, 69, 70} are being used in different 3D reconstruction applications. 
\subsection{Object Datasets}
\subsubsection{Pix3D and Sun RGB-D}
Pix3D \cite{59} and SUN RGB-D \cite{60} datasets are indoor scene objects datasets. These two datasets are good candidates for evaluating object quality and background of scene, respectively. Both datasets contain more than 10,000 frames along with respective 3D representations. Pix3D classified the objects into 9 different object categories and Sun RGB-D provides 3D room layout, 2D segmentation mask and 3D bounding box. 
\begin{table*}[]
\hspace{-2em}
\begin{tabular}{|l|l|l|l|l|}
\hline
\textbf{Authors} & \textbf{Dataset Type} & \textbf{Number of Subjects} & \textbf{Samples} & \textbf{Description} \\ \hline
Pix3D \cite{59} & Indoor Object Dataset & \begin{tabular}[c]{@{}l@{}}369 shapes of 9\\ categories\end{tabular} & 10,069 & \begin{tabular}[c]{@{}l@{}}This dataset enables \\ 3D shape modeling of \\ indoor objects.\end{tabular} \\ \hline
SUN RGB-D \cite{60} & Indoor Scene Dataset & \begin{tabular}[c]{@{}l@{}}47 Scenes \\ 800 Objects\end{tabular} & 10,335 & \begin{tabular}[c]{@{}l@{}}This dataset is very diverse for \\ construction of full scene including\\ all the objects along with their 2D\\ and 3D representation.\end{tabular} \\ \hline
\begin{tabular}[c]{@{}l@{}}Joint 2D-3D\\ Data \cite{61}\end{tabular} & Indoor Scene Dataset & \begin{tabular}[c]{@{}l@{}}270 scenes in \\ 11 categories\end{tabular} & 70,496 & \begin{tabular}[c]{@{}l@{}}Large scene dataset containing \\ depths, surface maps and \\ semantic annotations.\end{tabular} \\ \hline
Matterport3D \cite{62} & Indoor Scene Dataset & \begin{tabular}[c]{@{}l@{}}10,800 views of \\ 90 buildings\end{tabular} & 194,400 & \begin{tabular}[c]{@{}l@{}}Large Scale indoor scene \\ dataset having dept and semantic \\ segmentation. This dataset captures\\ view in 360 in the form of panoramic\\ images.\end{tabular} \\ \hline
Heritage-Recon \cite{62*} & \begin{tabular}[c]{@{}l@{}}Outdoor Structure\\ Dataset\end{tabular} & \begin{tabular}[c]{@{}l@{}}Four Landmarks\\ Structures\end{tabular} & - & \begin{tabular}[c]{@{}l@{}}This dataset is good for outdoor \\ \\ building reconstruction. This data\\ also contains point cloud of the\\ outer structure.\end{tabular} \\ \hline
CLOTH4D \cite{65} & Cloth Human Dataset & 1000 Humans & 100,000 & \begin{tabular}[c]{@{}l@{}}This data provides quite detailed\\ clothing texture along with \\ full human body can be used\\ 3D clothed human reconstruction.\end{tabular} \\ \hline
THuman \cite{49} & \begin{tabular}[c]{@{}l@{}}Single frame full body\\ human dataset\end{tabular} & \begin{tabular}[c]{@{}l@{}}7000 Human\\ meshes\end{tabular} & 28,000 & \begin{tabular}[c]{@{}l@{}}This data contains diverse action\\ dataset that can be utilized for \\ 3D reconstruction using single frame.\end{tabular} \\ \hline
SmartPortraits \cite{68} & \begin{tabular}[c]{@{}l@{}}Video based full body\\ human dataset\end{tabular} & 200 & 1000 (videos) & \begin{tabular}[c]{@{}l@{}}It is a video based dataset captured\\ from mobile phones in uncontrolled\\ condition. This can help in 3D human \\ reconstruction in the wild.\end{tabular} \\ \hline
HSiP-2 \cite{69} & \begin{tabular}[c]{@{}l@{}}Physical fitness 3D\\ action dataset\end{tabular} & 3 actions in 2 views & 80,440 & \begin{tabular}[c]{@{}l@{}}This data is domain specific and can be\\ used for intense exercise actions.\end{tabular} \\ \hline
MVP-Human dataset \cite{70} & \begin{tabular}[c]{@{}l@{}}Multi-frame Clothed\\ Human Dataset\end{tabular} & \begin{tabular}[c]{@{}l@{}}400 subjects with 15\\ poses and 8 views\end{tabular} & 48,000 & \begin{tabular}[c]{@{}l@{}}A good quality dataset for\\ clothed human with\\ multiple poses.\end{tabular} \\ \hline
\end{tabular}
\caption{\label{dataset-table}3D Object and Human Reconstruction Datasets}
\end{table*}
\subsubsection{Joint 2D-3D Data}

2D-3D dataset \cite{61} provides indoor scene images with 2D and 3D object representation by providing segmentation and geometric annotations. This dataset comprised 70,000 images captured in multiple indoor locations. Depth, surface normal, camera details and annotations are also provided in the dataset.

\subsubsection{Matterport3D}
In Matterport3D data \cite{62} is a large and diversified data captured in indoor environments. This dataset comprised 10,800 views in 194,400 RGB-D images. Each image is a panorama representation along with depth information, camera angles, 2D and 3D semantic segmentation.  Different properties of the dataset including global alignment and different views allows keypoint matching and view overlap. 

\subsubsection{Heritage-Recon}
Heritage-Recon \cite{62*} is a new benchmark dataset used to evaluate the in-the-wild surface sciences of outdoor structures. This data is a derived set from Open Heritage 3D and processed using structure from motion (SfM) to create point clouds. In this dataset four landmarks are primarily focused: Brandenburg Gate, Lincoln Memorial, Pantheon Exterior and Palacio de Bellas Artes. RGB images for each scene are also collected and using SfM camera position and point cloud is extracted.
\subsubsection{OmniObject3D}
This is a very recent and large dataset \cite{63} of 6000 scanned objects belong to 190 different daily life categories. Each of the object is captured using 2D and 3D input devices. Alongside images they provide 3D meshes, multi-view rendered images and point clouds. This data provide four evaluation benchmarks 1) 3D perception, 2) novel viewpoint 3) 3D object generation 4) surface reconstruction. 
\subsection{Human Datasets}
\subsubsection{CLOTH4D}
CLOTH4D \cite{65} provides a dataset of clothed humans for enabling the evaluation of clothed human reconstruction in a 3D environment. In this dataset 1000 subjects with different appearances and outfits are covered. They also provide 100,000 pairs of clothed meshes and unclothed humans. They provides quite detailed clothing texture and covered subjects in different motions with 289 actions.
\subsubsection{THuman}
Zheng et. al \cite{49} proposed THuman dataset to enable creation of 3D model from single image. This set contains 7000 single image based models. Each image is accompanied with respective 3D surface mesh, depth map and SMPL \cite{52} model. This dataset is not a synthesized dataset rather it provides 3D human data with surface geometry of humans with normal real life clothing. The authors utilized different view angles to produce nearly 28K images for training their proposed system. 
\subsubsection{3D Human Body Dataset}
Jakson et. al. \cite{67} refined the Human3.6M \cite{71} data to produce a new 3D reconstruction dataset. The produced meshes are voxelised to a dimensions of $128\times128\times128$. The details about facial expression and fingers are not fully covered in this dataset. The meshes are aligned with the mean of the z component that enabled suitable 3D structure regression without any issue. 

\subsubsection{SmartPortraits}
SmartPortraits \cite{68} is a very unique dataset containing video sequences of humans recorded in natural environments with uncontrolled settings from smartphones. There are overall 1000 videos in this dataset having 200 individuals captured in multiple different poses and locations. Depth of the respective image is taken from Kinect depth camera. The main purpose of this data is to fill the gap between raw mobile phone images and 3D reconstruction applications.
\subsubsection{$HSiPu^2$}
This dataset \cite{69} is a human physical fitness action dataset that is proposed for the purpose of action recognition and 3D reconstruction. This data is comprised of 80,440 images with three major actions: sit-up, push-up and pull-up.They utilized SMPL-X \cite{52} model and claimed that their dataset is good candidate for evaluation 3D action reconstruction. 
\subsubsection{MVP-Human dataset}
MVP-Human dataset \cite{70} is a multi-frame 3D reconstruction dataset that solve the problem of 3D clothed human to 3D clothed human avatar using multiple frames. This dataset is independent from the action, camera detail and environment. This set comprises 400 subjects captured in 15 different poses and 8 different view images are captured for each pose making the total number of 3D objects to 6000 with 48,000 images.

\section{Trade-Off Analysis}
Improving the 3D representation and quality can increase the computational cost, memory, and data requirements. Here we discuss three different trade-off analyses.
\subsection{Fidelity and Real-Time Capability}
Voxel- and implicit-function-based methods, including BodyNet, DeepHuman, and PIFu, produce high-quality geometry and mesh but require high computation during inference because of their dense sampling. This high demand of computation and time makes them unsuitable for real-time applications like virtual try-on systems. 
Gaussian Splatting-based 3D reconstruction methods support real-time rendering but sometimes compromise the fidelity in occluded regions. GaussianAvatar, 3DGS-Avatar, and LHM are examples of this type, where real-time response is achieved at the cost of quality. These types of systems are suitable for live gaming and AR/VR applications but are not suitable for garment design reconstruction and medical simulations, where accuracy is more important than inference time.   
\subsection{Identity Preservation and  Generalization} 
Methods that prioritize face identity preservation using face-based loss optimization, like GANFIT, preserve subject-specific details. These methods are mostly trained on face-centered datasets. However, these systems often struggle to generalize to full body and loose clothes. \\
Feed-forward Gaussian methods try to solve this trade-off by using a dedicated encoder to ensure identity. On the other hand, VLM-based methods use language-based feedback to improve both body generalization and facial identity. 
\subsection{Capture Constraints and Real-life Practicality}
Multi-view and depth-sensor-based systems capture surface details but require a fully controlled multi-view setup in a studio setting. This controlled environment can not be achieved outside the studio environment. On the other hand, single-image or monocular video-based methods sacrifice accuracy, especially in occluded regions, but remain a viable option for consumer-end applications.  
\section{Open Challenges}
\begin{figure}[t]
    \centering
    \includegraphics[width=0.85\linewidth]{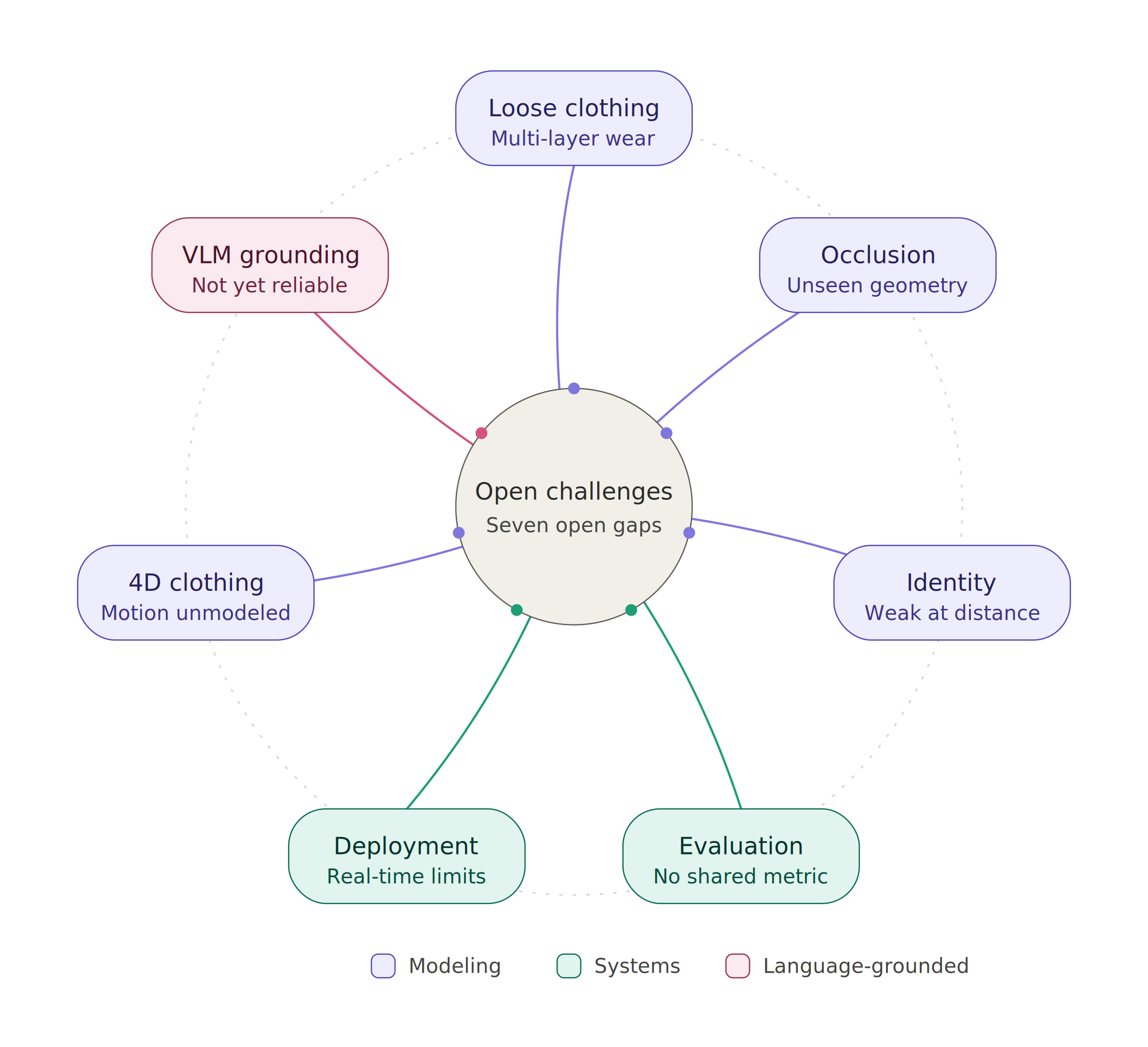}
    \caption{Open challenges in identity- and cloth-aware 3D human reconstruction, grouped 
    by category: modeling (purple), systems (teal), and language-grounded reliability (pink).}
    \label{fig:open_challenges}
\end{figure}
\subsection{Loose and Multi-Layer Clothing} 

Cloth-aware systems like SMPLicit and USR tend to treat clothing layers as a separate layer and body reconstruction as a fixed SMPL topology layer. This approach can fail when we have input subjects with loose clothing or multi-layered outfits. In this type of clothing, shape and style diverge considerably from the subject's body shape. Recent Gaussian splatting systems partially address the limitation, but a full clothing representation remains an open challenge.

\subsection{Unseen-Region Reconstruction and Occlusion}
Methods that depend on a single image are required to generate the full geometry and structure from only the visible part of the body. This involves predicting the texture and pose of body parts that are occluded or not visible in the camera. Regression-based methods try to produce over-smoothed meshes, such as diffusion-based systems produce more realistic poses but at the cost of identity. An optimal balance between plausibility and recognition is still a pending challenge.

\subsection{Identity Preservation for Complex Scenarios}

 Identity preservation using different types of losses works fine when the subject is near and well-lit. However, identity degrades in dim lighting, extreme poses, and for distant subjects. LHM works on improving the identity by adding a dedicated encoder for this task but a model that can preserve the identity for different complex scenarios is still an unresolved challenge.

\subsection{Real-Time and On-Device Deployment}
Consumer end applications mostly require processing on edge devices. Most high-fidelity systems require a GPU-based, high-end machine to run and can take a few seconds to generate a realistic 3D mesh. Gaussian Splatting methods tried to resolve the issue of computational complexity, but personalization of an avatar for a new subject still generally requires a high-end system.

\subsection{4D Clothing Behavior}
Real-world cloth movements, styles, wrinkles, and motion-dependent deformation are not clearly captured by the current systems. However, 4D-DRESS is a recently proposed 4D dataset to allow researchers to work in this direction.

\subsection{Reliability of Language-Grounded Supervision} 
When should we trust the language-guiding signals is an open question for VLM-based agentic verification loops.  Textual constraints generated by VLM mainly depend on VLM visual grounding. If a VLM visual grounding fails for an occluded or unoccluded region, it can hallucinate the textual constraint and can misguide the learning or reconstruction process. This failure mode is only associated with language-based methods, as compared to the previous geometric reconstruction methods.

VLM-generated textual constraints and agentic verification loops are only as reliable as the underlying VLM's own visual grounding. A VLM that misreads occluded or ambiguous body parts could introduce hallucinated constraints that steer reconstruction toward a confidently wrong result, a distinct failure mode from the purely geometric ambiguities discussed in 6.1–6.3. Establishing when to trust language-grounded feedback over, or combine it with, purely geometric signals remains an open question in this emerging direction.

\section{Discussion}
Overall, researchers in 3D reconstruction have focused on improving various aspects, including performance time, realism, and generalization. Researchers also worked on simplifying the requirements of capturing systems by enabling real-time reconstruction with simple camera. \\
The traditional 3D human reconstruction systems are predominantly developed around high freedom capturing systems that provide more information alongside 2D image. Either by employing multiple cameras in different locations for capturing multi-view stream \cite{54, 76} or by using depth sensing capturing systems like depth cameras and kinetics \cite{75}. The researchers accomplished high quality reconstruction by splitting the process into two step process, initially construct multi-view stereo followed by merging depth maps of different view-points. Nevertheless, the overall time required for processing the data gathered from different input sources is very high that makes it impractical for real-life applications. These depth sensing and multi-view cameras are limited by the range distance of the sensors and controlled setting of cameras. These limitations makes these work unfit for real-life applications. \\

There are different types of learning-based 3D reconstruction systems including voxels and meshes. A large set of studies focuses on methods based on SMPL \cite{52}. SMPL was developed by Loper et al. in 2015 \cite{52}, and is a statistical model for the reconstructing human body. It can accommodate different shapes and poses of diverse subjects through parameter modification. Typically, all methods that are based on SMPL utilize it as a geometric foundation. They generated body meshes from images,  leveraging the low-dimensional nature of the SMPL model. Simple SMPL-based methods are not fully refined and are not of high quality.\\

Voxels-based and meshes-based methods are refined and produce high-quality reconstruction surfaces but they require large memory for storage of large numbers of voxels. It is not feasible for light weight applications. To cater this high memory demand, researchers have been proposing implicit functions that directly learn 2D to 3D mapping using \cite{77}. These types of functions and networks require a large annotated data to learn 2D to 3D mapping. Furthermore, ground-truth based learning limits the model in terms of generalization as model learns dataset dependent features.\\

Some recent attempts have been made to learn 3D model representation without depending on the annotated datasets. These methods utilize Neural learning to learn rendering for 3D reconstruction. These networks can learn voxels, point cloud, texture meshes, and implicit functions. A novel view construction is proposed by neural radiance field (NeRF) \cite{78}. Similarly, Peng et al. \cite{79, 80} utilized the power of NeRF in combination with SMPL for dynamic human 3D reconstruction.

\section{Conclusion}
In conclusion, this survey enlists different image-based deep and machine learning methods for single and multi-view 2D to 3D Human Reconstruction. It initially groups different prerequisite high-level features (age, gender, pose, body skeleton, and weight) for  generating their identifiable and relatable 3D model.\\
We also reviewed different 3D object reconstruction systems of human objects along with their underlying methodologies, training paradigms, datasets and performance results. Reviewed 3D human reconstruction systems are classified into simple, identity-aware and cloth-aware reconstructions. Finally, we also present a comprehensive dataset review including human and non-human objects reconstruction datasets. We also provided a comparison table for showing the statistics of these reviewed datasets.

 \section*{Conflict of interest}
The authors declare that they have no conflict of interest.

\bibliographystyle{unsrt}     
\bibliography{references}   

\label{refer}

\end{document}